\documentclass[12pt,runningheads]{llncs}
\usepackage{graphicx}
\usepackage{amsfonts, amsmath, amsxtra, amssymb, latexsym,url}
\usepackage{makeidx}
\urldef{\mailsa}\path{berikov@math.nsc.ru}
\newcommand{\keywords}[1]{\par\addvspace\baselineskip
\noindent\keywordname\enspace\ignorespaces#1}
\usepackage[cp1251]{inputenc}
\usepackage[english]{babel}
\usepackage{multirow}
\def\H{\mathcal{H}}

\usepackage{color}

\newcommand{\bC}{W}
\newcommand{\bL}{{U}}
\newcommand{\bI}{{I}}

\usepackage{geometry}
\begin{document}

\pagestyle{headings}

\title{Semi-Supervised Regression using Cluster Ensemble and
Low-Rank Co-Association Matrix Decomposition under Uncertainties}
\titlerunning{Semi-Supervised Regression using Cluster Ensemble}
\author{Vladimir Berikov\inst{1,2}\orcidID{0000-0002-5207-9764} \and Alexander Litvinenko\inst{3} \orcidID{0000-0001-5427-3598}}

\authorrunning{V. Berikov \and A. Litvinenko}

\institute{Sobolev Institute of mathematics,
Novosibirsk, Russia\\ \and
Novosibirsk State University, Novosibirsk, Russia,
\email{berikov@math.nsc.ru}\\
\url{http://www.math.nsc.ru}
\and RWTH Aachen, Germany,
\url{www.uq.rwth-aachen.de}\\
\email{litvinenko@uq.rwth-aachen.de}}



\maketitle

\begin{abstract}
In this paper, we solve a semi-supervised regression
problem. Due to the lack of knowledge about the
data structure and the presence of random noise, the considered data
model is uncertain. We propose a  method which combines graph
Laplacian regularization and cluster ensemble methodologies. The
co-association matrix of the ensemble is calculated on both labeled
and unlabeled data; this matrix is used as a similarity matrix in
the regularization framework to derive the predicted outputs. We use
the low-rank decomposition of the co-association matrix to
significantly speedup calculations and reduce memory.
Numerical experiments using the Monte Carlo
approach demonstrate robustness, efficiency, and scalability of the
proposed method.
\end{abstract}



\section{Introduction}
Machine learning problems can be classified as supervised,
unsupervised and semi-supervised. Let data set $ \mathbf{X} =
\{x_{1},\dots,x_{n}\}$ be given, where $x_{i} \in \mathbb{R}^d$ is
feature vector, $d$ is feature space dimensionality. In a supervised
learning context, we are given an additional set
$Y=\{y_1,\dots,y_{n}\}$ of target feature values (labels) for all
data points, $y_i \in D_Y$, where $D_Y$ is target feature domain. In the case of supervised classification, the domain is an unordered set of
categorical values (classes, patterns). In case of supervised
regression, the domain $D_Y \subseteq \mathbb{R}$. Using this
information (which can be thought as provided by a certain
``teacher''), it is necessary to find a decision function $y=f(x)$
(classifier, regression model) for predicting target feature values
for any new data point $x \in \mathbb{R}^d$ from the same
statistical population. The function should be optimal in some
sense, e.g., give minimal value to the expected losses.

In an unsupervised learning setting, the target feature values are not
provided. The problem of cluster analysis, which is an important
direction in unsupervised learning, consists in finding a partition
$P = \{C_1 ,\dots,C_K \}$ of $\mathbf{X}$ on a relatively small
number of homogeneous clusters describing the structure of data. As
a criterion of homogeneity, it is possible to use a functional
dependent on the scatter of observations within groups and the
distances between clusters. The desired number of clusters is either
a predefined parameter or should be found in the best way.

We note that since the final cluster partition is uncertain due to random noise in sample data, lack of knowledge about the feature set and the data structure, parameters, weights, and  initialization settings, a set of different cluster partitions is calculated. Then a final cluster partition is formed.

In semi-supervised learning problems, the target feature values are
known only for a part of data set $X_1\subset \mathbf{X}$. It is
possible to assume that $X_1=\{x_{1},\dots,x_{n_1}\}$, and the
unlabeled part is $X_0=\{x_{n_1+1},\dots,x_{n}\}$. The set of labels
for points from $X_1$ is denoted by $Y_1=\{y_1,\dots,y_{n_1}\}$. It
is required to predict target feature values as accurately as
possible either for given unlabeled data $X_0$ (i.e., perform
\emph{transductive learning}) or for arbitrary new observations from
the same statistical population (\emph{inductive learning}). In
dependence of the type of the target feature, one may consider
semi-supervised classification or semi-supervised regression
problems \cite{sslsurvey}.

The task of semi-supervised learning is important because in many
real-life problems only a small part of available data can be
labeled due to the large cost of target feature registration. For
example, manual annotation of digital images is rather time-consuming. Therefore labels can be attributed to only a small part
of pixels. To improve prediction accuracy, it is necessary to use
information contained in both labeled and unlabeled data. An
important application is hyperspectral image semi-supervised
classification \cite{Camps-Valls}. 

In this paper, we consider a semi-supervised regression problem in the
transductive learning setting. In semi-supervised regression, the
following types of methods can be found in the literature
\cite{Kostopoulos}: co-training \cite{ZhouZ}, semi-supervised kernel
regression \cite{Wang}, graph-based and spectral regression methods
\cite{Wu,Doquire,Zhao}, etc.

We propose a novel semi-supervised regression method using a
combination of graph Laplacian regularization technique and cluster
ensemble methodology. Graph regularization (sometimes called
manifold regularization) is based on the assumption which states
that if two data points are on the same manifold, then their
corresponding labels are close to each other. A graph Laplacian  is
used  to measure the smoothness of the predictions on the data
manifold including both labeled and unlabeled data
\cite{ZhouD,Belkin}.

Ensemble clustering aims at finding consensus partition of data
using some base clustering algorithms. As a rule, application
of this methodology allows one to get a robust and effective solution,
especially in case of uncertainty in the data model. Properly organized
ensemble (even composed of ''weak'' learners) significantly improves
the overall clustering quality \cite{Boongoen}.

Different schemes for applying ensemble clustering for
semi-supervised classification were proposed in \cite{Yu,BerikovS}.
The suggested methods are based on the hypothesis which states that
a preliminary ensemble allows one to restore more accurately metric
relations in data in noise conditions. The obtained co-association
matrix (CM) depends on the outputs of clustering algorithms and is
less noise-addicted than a conventional similarity matrix. This
increases the prediction quality of the methods. 

The same idea is pivotal in the proposed semi-supervised regression
method. We assume a statistical connection between the clustering
structure of data and the predicted target feature. Such a
connection may exist, for example, when some hidden classes are
present in data, and the belonging of objects to the same class
influences the proximity of their responses.

To decrease the computational cost and the storage requirement and
to increase the scalability of the method, we suggest usage of
low-rank (or hierarchical) decomposition of CM. This decomposition
will reduce the numerical cost and storage from cubic to
(log-)linear \cite{HackHMEng}.

Parametric  approximations,  given  by  generalized  linear  models, as well as nonlinear models, given by neural networks were compared in \cite{Law_Litv18}.

In the rest of the paper, we describe the details of the suggested
method. Numerical experiments are presented in the correspondent
section. Finally, we give concluding remarks.

\section{Combined semi-supervised regression and ensemble clustering}
\label{sec:operations}
\subsection{Graph Laplacian regularization}
\label{sec:graph}
We consider a variant of graph Laplacian regularization in
semi-supervise transductive regression which solves the following
optimization problem:

find $f^{*}$ such that $f^{*}= \arg\min\limits_{f\in \mathbb{R}^n} Q(f)$, where \\
\begin{equation}
\label{eq1}
Q(f):=\frac{1}{2} \left( \sum\limits_{x_i \in X_1}
(f_i-y_i)^2 +
\alpha \sum\limits_{x_i,x_j \in \mathbf{X}}w_{ij}(f_i-f_j)^2 +\beta
||f||^2   \right),
\end{equation}

\noindent $f=(f_1,\dots,f_n)$ is a vector of predicted outputs: $f_i=f(x_i)$;
$\alpha, \beta>0$ are regularization parameters, $W=(w_{ij})$ is
data similarity matrix. The degree of similarity between points
$x_i$ and $x_j$ can be calculated using appropriate function, for
example from the Mat\'{e}rn family \cite{Matern1986a}. The
Mat\'{e}rn function depends only on the distance $h:=\Vert
x_i-x_j\Vert $ and is defined as
$W(h)=\frac{\sigma^2}{2^{\nu-1}\Gamma(\nu)}\left(\frac{h}{\ell}\right)^\nu K_\nu\left(\frac{h}{\ell}\right)$
with three parameters $\ell$, $\nu$, and $\sigma^2$. For instance,
$\nu=1/2$ gives the well-known exponential kernel
$W(h)=\sigma^2\exp(-h/\ell)$, and $\nu=\infty$ gives the Gaussian
kernel $W(h)=\sigma^2\exp(-h^2/2\ell^2)$.

In this paper we also use RBF kernel with parameter $\ell$:
$w_{ij}=\exp(-\frac{\Vert x_i-x_j\Vert^2}{2 \ell^2})$.

The first term in right part of (\ref{eq1}) minimizes fitting error
on labeled data; the second term aims to obtain ''smooth''
predictions on both labeled and unlabeled sample; the third one is
Tikhonov's regularizer.

Let graph Laplacian be  denoted by $L=D-W$ where $D$ be a diagonal
matrix defined by $D_{ii} = \sum\limits_{j} w_{ij}$. It is easy to
show (see, e.g., \cite{Belkin,ZhouD}) that 
\begin{equation}
\label{eq:propL}
 \sum\limits_{x_i,x_j \in
\mathbf{X}}w_{ij}(f_i-f_j)^2=2f^T L f.
\end{equation}

Let us introduce vector $Y_{1,0}=(y_1,\dots,y_{n_1},
\underbrace{0,\dots,0}_{n-n_1})^T$, and let $G$ be a diagonal
matrix:
\begin{equation}
\label{G} G=diag(G_{11}\dots,G_{nn}), \; G_{ii} =\left\{^{\beta+1,
\; i=1,\dots,n_1}_{\beta, \; i=n_1+1,\dots,n,} \right. .
\end{equation}

Differentiating $Q(f)$ with respect to $f$, we get
$$
\frac{\partial Q}{\partial f}\mid_{f=f^{*}}=Gf^{*}+\alpha
Lf^{*}-Y_{1,0}=0,
$$
hence
\begin{equation}
\label{f*} f^{*}=(G+\alpha L)^{-1}\;Y_{1,0}
\end{equation}
under the condition that the inverse of matrix sum exists (note that
the regularization parameters $\alpha,\beta$ can be selected to
guaranty the well-posedness of the problem). Numerical methods such
as Tikhonov or Lavrentiev regularization \cite{Tikhonov} can also be
used to obtain the predictions.

\subsection{Co-association matrix of cluster ensemble}
\label{sec:Co-ass}
In the proposed method, we use a co-association matrix of cluster
ensemble as similarity matrix in (\ref{eq1}). Co-association matrix
is calculated as a preliminary step in the process of cluster
ensemble design with various clustering algorithms or under
variation across a given algorithm's parameter settings \cite{Fred}.

Let us consider a set of partition variants $\{P_{l}\}_{l=1}^r$,
where $P_l = \{C_{l,1} ,\dots,C_{l,K_l} \}$, $C_{l,k} \subset
\mathbf{X}$, $C_{l,k}\bigcap C_{l,k'}=\varnothing$,  $K_l$ is number
of clusters in $l$th partition. For each $P_l$ we determine matrix
$H_l=(h_l(i,j))_{i,j=1}^n$ with elements indicating whether a pair
$x_i$, $x_j$ belong to the same cluster in $l$th variant or not:
$h_l(i,j)=\mathbb{I}[c_l(x_{i} ) = c_l(x_{j} )]$, where $\mathbb{I}(
\cdot )$ is indicator function ($\mathbb{I}[true]=1$,
$\mathbb{I}[false]=0$), $c_l(x)$ is cluster label assigned to $x$.
The weighted averaged co-association matrix (WACM) is defined as
follows: 
\begin{equation}
\label{eq:defH}
H = (H(i,j))_{i,j=1}^n, \quad H(i,j) = \sum\limits_{l=1}^r
w_l H_l(i,j)
\end{equation}
where $w_1,\dots,w_r$ are weights of ensemble
elements, $w_l \ge 0$, $\sum w_l=1$. The weights  should reflect the
``importance'' of base clustering variants in the ensemble \cite{Berikov2003} and be
dependent on some evaluation function $\Gamma$ (cluster validity
index, diversity measure) \cite{Berikov2017}: $w_l=\gamma_l /
\sum\limits_{l'} \gamma_{l'}$, where $\gamma_l=\Gamma(l)$ is an
estimate of clustering quality for the $l$th partition (we assume
that a larger value of $\Gamma$ manifests better quality).

In the methodology presented in this paper, the elements of WACM are
viewed as similarity measures learned by the ensemble. In a sense,
the matrix specifies the similarity between objects in a new feature
space obtained utilizing some implicit transformation of the
initial data. The following property of WACM allows increasing the
processing speed.

\noindent\emph{Proposition 1.} Weighted averaged co-association
matrix admits low-rank decomposition in the form:
\begin{equation}
\label{H} H = B B^T, \;B=[B_1 B_2 \dots B_r]
\end{equation}
where $B$ is a block matrix, $B_l=\sqrt{w_l}\, A_l$, $A_l$ is ($n
\times K_l$) cluster assignment matrix for $l$th partition:
$A_l(i,k)=\mathbb{I}[c(x_i)=k]$, $i=1,\dots,n$, $k=1,\dots,K_l$.

The proof is fairly straightforward and is omitted here for the
sake of brevity. As a rule, $m=\sum_l K_l \ll n$, thus (\ref{H})
gives us an opportunity of saving memory by storing $(n \times m)$
sparse matrix instead of full $(n \times n)$ co-association matrix.
The complexity of matrix-vector multiplication $H \cdot x$ is
decreased from $O(n^2)$ to $O(nm)$.
\subsection{Cluster ensemble and graph Laplacian regularization}
\label{sec:ClAns}
Let us consider graph Laplacian in the form: $L'=D'-H$, where
$D'=\text{diag}(D'_{11},$ $\dots, D'_{nn})$, $D'_{ii} =
\sum\limits_{j} H(i,j)$. We have:

\begin{multline}\small
\label{D'} D'_{ii} = \sum\limits_{j=1}^{n} \sum\limits_{l=1}^{r} w_l
\sum\limits_{k=1}^{K_l} A_l(i,k) A_l(j,k) = \\
 \sum\limits_{l=1}^{r}
w_l \sum\limits_{k=1}^{K_l} A_l(i,k) \sum\limits_{j=1}^{n}A_l(j,k)
=\sum\limits_{l=1}^{r} w_l N_{l}(i)
\end{multline}
\noindent where $N_{l}(i)$ is the size of the cluster which includes
point $x_i$ in $l$th partition variant.

Substituting $L'$ in (\ref{f*}), we obtain cluster ensemble based
predictions of output feature in semi-supervised regression:
\begin{equation} \label{f**}
f^{**}=(G+\alpha L')^{-1}\;Y_{1,0}.
\end{equation}
Using a low-rank representation of $H$, this expression can be
transformed into the form which involves more efficient matrix
operations:
$$f^{**}=(G+\alpha D'-\alpha B
B^T)^{-1}\;Y_{1,0}.$$ In linear algebra, the following Woodbury
matrix identity is known:
\begin{equation*}
(S + UV)^{-1} = S^{-1} - S^{-1} U(I + V S^{-1} U)^{-1} VS^{-1},
\end{equation*}
where $S \in \mathbb{R}^{n\times n}$ is invertible matrix, $U \in
\mathbb{R}^{n\times m}$ and $V \in \mathbb{R}^{m\times n}$. We can
denote $S=G+\alpha D'$ and get
\begin{equation} \label{S}
S^{-1}=\text{diag}(1/(G_{11}+\alpha D'_{11}), \dots,1/(G_{nn}+\alpha
D'_{nn})),
\end{equation}
where $G_{ii}, D'_{ii}$, $i=1,\dots,n$ are defined in (\ref{G}) and
(\ref{D'}) correspondingly.

Now it is clear that the following statement is valid:

\noindent\emph{Proposition 2.} Cluster ensemble based target feature
prediction vector (\ref{f**}) can be calculated using low-rank
decomposition as follows:
\begin{equation} \label{f**lr}
f^{**}=(S^{-1}+\alpha S^{-1}B (I-\alpha B^T
S^{-1}B)^{-1}BS^{-1})\;Y_{1,0}
\end{equation}
where matrix $B$ is defined in (\ref{H}) and $S^{-1}$ in (\ref{S}).

Note that in (\ref{f**lr}) we need to invert significantly smaller
($m \times m$) sized matrix instead of ($n \times n$) in (\ref{f**}).
The overall computational complexity of (\ref{f**lr}) can be
estimated as $O(nm+m^3)$.

The outline of the suggested algorithm of semi-supervised regression
based on the low-rank decomposition of the co-association matrix (SSR-LRCM)
is as follows.

\bigskip

\noindent\textbf{Algorithm SSR-LRCM}

\noindent\textbf{Input}:

\noindent ${ {\mathbf X}}$: dataset including both labeled and
unlabeled sample;

\noindent $Y_1$: target feature values for labeled instances;

\noindent $r$: number of runs for base clustering algorithm $\mu$;

\noindent  $\mathbf{\Omega}$: set of parameters (working conditions)
of clustering algorithm.

\noindent\textbf{Output}:

\noindent $f^{**}$: predictions of target feature for labeled and
unlabeled objects.

\noindent\textbf{Steps:}

\noindent 1. Generate $r$ variants of clustering partition with
algorithm $\mu$ for working parameters randomly chosen from
$\mathbf{\Omega}$; calculate weights $w_1,\dots,w_r$ of variants.

\noindent 2. Find graph Laplacian in low-rank representation using
matrices $B$ in (\ref{H}) and $D'$ in (\ref{D'});

\noindent 3. Calculate predictions of target feature according to
(\ref{f**lr}).

\noindent\textbf{end.}

In the implementation of SSR-LRCM, we use K-means as base clustering
algorithm which has linear complexity with respect to data
dimensions.

\section{Hierarchical Approximation}
\label{sec:Hcov} In this section we discuss the case if matrices $W$
and $H$ do not have any low-rank decomposition or this low-rank is
expensive (e.g., the rank is comparable with $n$). In that case then
one can try to apply, so-called, hierarchical matrices
($\H$-matrices), introduced in \cite{Part1}, \cite{HackHMEng}. 

The $\mathcal{H}$-matrix format has a log-linear computational cost\footnote{log-linear means $\mathcal{O}(kn\log n)$, where the rank $k$ is a small integer, and $n$ is the size of the data set} and storage.
The $\mathcal{H}$-matrix technique allows us to efficiently work with general matrices $W$ and $H$  (and not only with structured ones like Toeplitz, circulant or three diagonal). 
Another advantage is that all linear algebra operations from
Sections~\ref{sec:graph} and \ref{sec:Co-ass} preserve (or only slightly increase) the rank $k$ inside of each sub-block.

There are many implementations of $\H$-matrices exist, e.g., the HLIB library (http://www.hlib.org/), $\H^2$-library (https://github.com/H2Lib), and HLIBpro library (https://www.hlibpro.com/).
We used the HLIBpro library, which is actively supported commercial, robust, parallel, very tuned, and well tested library. 
Applications of the $\H$-matrix technique to the graph Laplacian can be found in the HLIBpro library\footnote{https://www.hlibpro.com/}, and to covariance matrices in \cite{khoromskij2009application} and in \cite{litvHLIBPro17}.

\begin{figure}[htbp!]
 \centering
       \includegraphics[width=5cm]{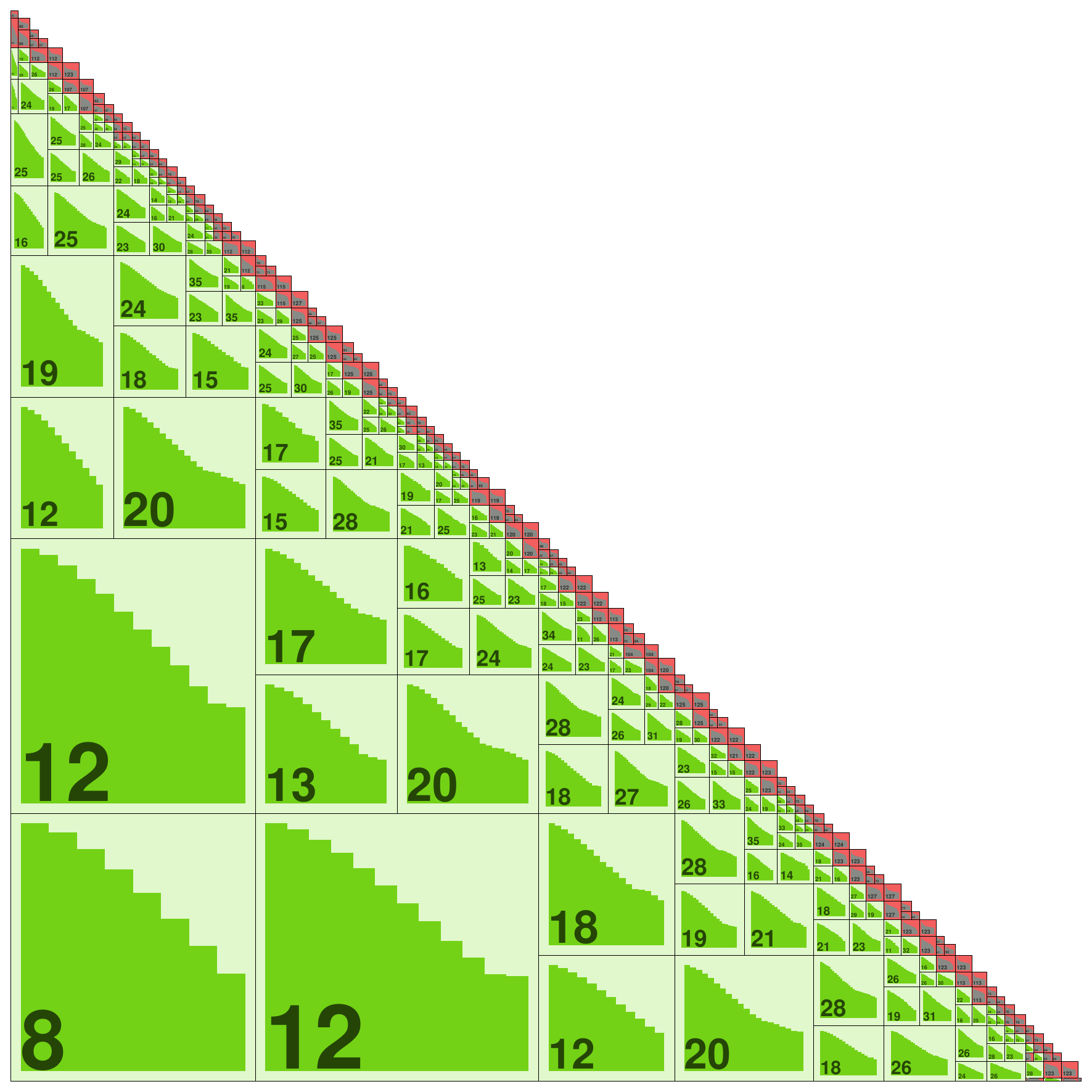}\vspace{-.1cm}
       \includegraphics[width=5cm]{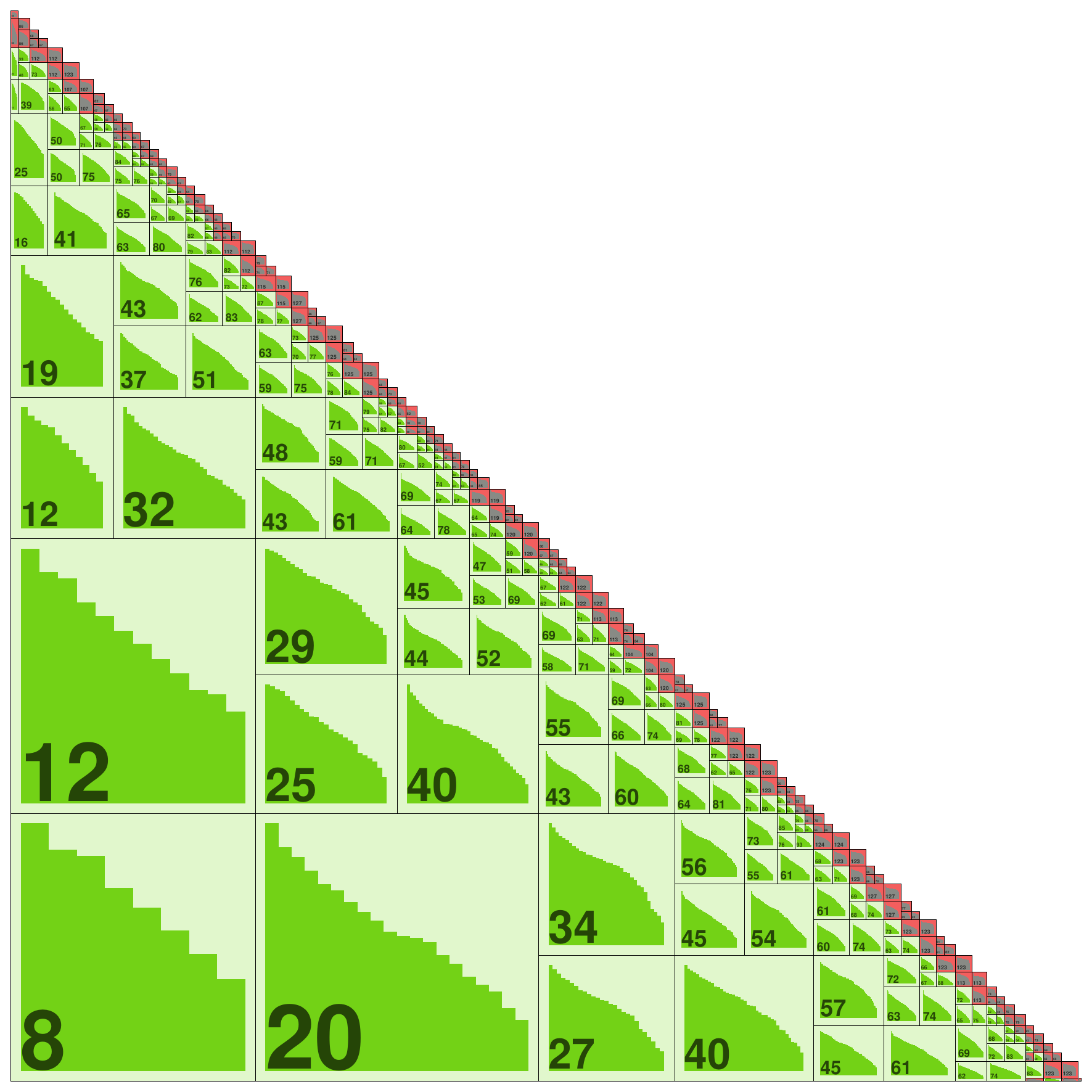}\vspace{-.1cm}
     \caption{(left) An example of the $\mathcal{H}$-matrix approximation $\widetilde{W}$ of an $n\times n$ matrix $W$, $n=16000$. (right) The corresponding Cholesky factor $\widetilde{U}$, where $\widetilde{W}=\widetilde{U}\widetilde{U}^\top$.}
 \label{fig:Hexample3}
 \end{figure}

The $\H$-matrix technique is defined as a hierarchical partitioning of a given matrix
into sub-blocks followed by the further approximation of the majority of these sub-blocks by low-rank matrices. Figure~\ref{fig:Hexample3} shows an example of the $\mathcal{H}$-matrix approximation $\widetilde{W}$ of an $n\times n$ matrix $W$, $n=16000$ and its Cholesky factor $\widetilde{U}$, where $\widetilde{W}=\widetilde{U}\widetilde{U}^\top$. The dark (or red) blocks indicate the dense matrices and the grey (green) blocks indicate the rank-$k$ matrices; the number inside each block is its rank. The steps inside the blocks show the decay of the singular values in $\log$ scale. 
The Cholesky factorization is needed for computing the inverse, $\widetilde{W}^{-1}=(\widetilde{U}\widetilde{U}^\top)^{-1}=\widetilde{U}^{-\top}\widetilde{U}^{-1}$. This way is cheaper as computing the inverse directly.

To define which sub-blocks can be approximated well by low-rank matrices and which cannot, a so-called admissibility condition is used (see more details in \cite{litvHLIBPro17}). There are different admissibility conditions possible: weak, strong, domain decomposition  based. Each one results in a new subblock partitioning.
Blocks that satisfy the admissibility condition can be approximated by low-rank matrices; see \cite{Part1}.


On the first step, the matrix is divided into four sub-blocks. 
Then each (or some) sub-block(s) is (are) divided again and again hierarchically until sub-blocks are sufficiently small.  The procedure stops when either one of the sub-block sizes is $n_{\mbox{min}}$ or smaller (typically $n_{\mbox{min}}\leq 128$), or when this sub-block can be approximated by a low-rank matrix.

Another important question is how
to compute these low-rank approximations. One (heuristic) possibility is the Adaptive Cross Approximation (ACA) algorithm \cite{HackHMEng}, which performs the approximations with a linear complexity $\mathcal{O}(kn)$ in contrast to $\mathcal{O}(n^3)$ by the standard singular value decomposition (SVD).


The storage requirement of $\widetilde \bC$ and the matrix vector multiplication cost $\mathcal{O}(kn\log n)$, the matrix-matrix addition costs $\mathcal{O}(k^2n\log n)$, and the matrix-matrix product and the matrix inverse cost $\mathcal{O}(k^2n\log^2 n)$; see \cite{Part1}.
In Table~\ref{table:approx_compare_rank} we show dependence of the two matrix errors on the $\H$-matrix rank $k$ for the Mat\'{e}rn function with parameters $\ell=\{0.25, 0.75\}$, $\nu=1.5$, and $x_i,x_j\in [0,1]^2$. 
We can bound the relative error $\Vert \bC^{-1} - \widetilde{\bC}^{-1}\Vert/\Vert \bC^{-1} \Vert$ for the approximation of the inverse as
\begin{equation*}
\frac{\Vert \bC^{-1} - \widetilde{\bC}^{-1}\Vert}{\Vert \bC^{-1} \Vert}=\frac{\Vert (\bI-  \widetilde{\bC}^{-1}\bC)\bC^{-1}\Vert}{\Vert \bC^{-1} \Vert} \leq \Vert (\bI-  \widetilde{\bC}^{-1}\bC)\Vert.
\end{equation*} 
$\Vert (\bI-  \widetilde{\bC}^{-1}\bC)\Vert_2$ can be estimated by few steps of the power iteration method.
The rank $k\leq 20$ is not sufficient to approximate
the inverse.
The spectral norms of $\tilde{\bC}$ are $\Vert \widetilde{\bC}_{(\ell=0.25)}\Vert_2=720$ and $\Vert \widetilde{\bC}_{(\ell=0.75)}\Vert_2=1068$.

\begin{table}[h!]
\centering
\caption{Convergence of the $\H$-matrix approximation error vs. the $\H$-matrix rank $k$ of a Mat\'{e}rn function with parameters $\ell=\{0.25, 0.75\}$, $\nu=1.5$, $x_i,x_j\in [0,1]^2$, $n=16{,}641$, see more in \cite{{LitvGentonSunKeyes17}}}
\begin{tabular}{|c|cc|cc|}
\hline
 $k$ & \multicolumn{2}{c|}{$\Vert \bC - \widetilde{\bC} \Vert_2$} & \multicolumn{2}{c|}{$\Vert \bI - \widetilde{\bC}^{-1}{\bC}  \Vert_2  $} \\
     &   $\ell=0.25$   &  $\ell=0.75$ &  $\ell=0.25$   &  $\ell=0.75$ \\
\hline
 20  &   5.3e-7& 2e-7   & 4.5& 72\\
 30  &   1.3e-9& 5e-10  & 4.8e-3& 20\\
 40  &  1.5e-11& 8e-12  & 7.4e-6& 0.5\\
 50  &  2.0e-13& 1.5e-13&  1.5e-7& 0.1\\
\hline
\end{tabular}
\label{table:approx_compare_rank}
\end{table}
Table~\ref{table:approx_W} shows the computational time and storage for the $\H$-matrix approximations \cite{LitvGentonSunKeyes17,litvHLIBPro17}. These computations are done with the parallel $\H$-matrix toolbox, HLIBpro. The number of computing cores is 40, the RAM memory 128GB.
It is important to note that the computing time (columns 2 and 5) and the storage cost (columns 3 and 6) are growing nearly linearly with $n$. Additionally, we provide the accuracy of the $\H$-Cholesky inverse.

\begin{table}[htbp!]
\centering
\caption{Computing times and storage costs of $\widetilde{W}\in \mathbb{R}^{n\times n}$. Accuracy in each sub-block is $\varepsilon=10^{-7}$.}
\begin{small}
\begin{tabular}{|c|ccc|ccc|}
\hline
$ n$ & \multicolumn{3}{c|}{$\widetilde{\bC}$} &  \multicolumn{3}{c|}{$\widetilde{\bL}\widetilde{ \bL}^\top$} \\
             & time & size & kB/$n$  & time & size &  $\Vert \bI-(\widetilde{\bL}\widetilde{\bL}^\top)^{-1}{\bC} \Vert_2$     \\
        &  sec & MB &     & sec &  MB &      \\
\hline
128{,}000      & 7.7 & 1160   & 9.5 & 36.7 & 1310 & $3.8\cdot 10^{-5}$\\
256{,}000      & 13 & 2550   & 10.5 & 64.0 & 2960 & $7.1\cdot 10^{-5}$ \\ 
512{,}000      & 23 & 4740   & 9.7 & 128 & 5800 & $7.1\cdot 10^{-4}$  \\ 
1{,}000{,}000 & 53 & 11260   & 11 & 361 & 13910 & $3.0\cdot 10^{-4}$ \\ 
2{,}000{,}000 & 124 & 23650   & 12.4 & 1001 & 29610 & $5.2\cdot 10^{-4}$ \\
\hline
\end{tabular}
\end{small}
\label{table:approx_W}
\end{table}
\subsection{$\H$-matrix approximation of regularized graph Laplacian}
\label{sec:Hgraph}
We rewrite formulas from Sections~\ref{sec:graph} - \ref{sec:ClAns} in  the $\H$-matrix format.
Let $\tilde{W}$ be an $\H$-matrix approximation of $W$. 
The new optimization problem will be:

find $\tilde{f}^{*}$ such that $\tilde{f}^{*}= \arg\min\limits_{f\in \mathbb{R}^n} \tilde{Q}(f)$, where \\
\begin{equation}
\label{eq1B}
\tilde{Q}(f):=\frac{1}{2} \left( \sum\limits_{x_i \in X_1}
(f_i-y_i)^2 +
\alpha \sum\limits_{x_i,x_j \in \mathbf{X}}\tilde{w}_{ij}(f_i-f_j)^2 +\beta
\Vert f \Vert ^2   \right).
\end{equation}
Using (\ref{eq:propL}) and assuming that the $\H$-matrix approximation error $\Vert \tilde{L} - L \Vert\leq \varepsilon$, obtain
\begin{equation}
\label{eq1:err}
\Vert \tilde{Q}(f) -Q(f) \Vert \leq \alpha \left( f^\top \tilde{L} f - f^\top {L} f\right)\leq \alpha \Vert f\Vert^2\Vert \tilde{L} - L \Vert=\Vert f\Vert^2 \varepsilon.
\end{equation}


Let the approximate graph Laplacian be  denoted by $\tilde{L}=\tilde{D}-\tilde{W}$ where $\tilde{D}$ be a diagonal
matrix defined by $\tilde{{D}}_{ii} = \sum\limits_{j} \tilde{w}_{ij}$.
Differentiating $\tilde{Q}(f)$ with respect to $f$, we get
$$
\frac{\partial \tilde{Q}}{\partial f}\mid_{f=\tilde{f}^{*}}={G}\tilde{f}^{*}+\alpha
\tilde{L}\tilde{f}^{*}-Y_{1,0}=0,
$$
hence
\begin{equation}
\label{Hf*} \tilde{f}^{*}=({G}+\alpha \tilde{L})^{-1}\;Y_{1,0}
\end{equation}
{The impact of the $\H$-matrix approximation error could be measured as follows 
\begin{equation}
\label{eq1B:err}
\Vert \tilde{f}^* - f^* \Vert \leq \Vert ({G}+\alpha \tilde{L})^{-1}- ({G}+\alpha {L})^{-1}\Vert \cdot \Vert Y_{1,0} \Vert
\end{equation}
or
\begin{equation}
\label{eq2:err}
\Vert \tilde{f}^* - f^* \Vert \leq \Vert (I+\alpha G^{-1}\tilde{L})^{-1}- ({I}+\alpha G^{-1}{L})^{-1}\Vert \Vert G \Vert \cdot \Vert Y_{1,0} \Vert
\end{equation}
Now, if matrix norm (e.g., spectral norm) of $\alpha G^{-1}\tilde{L}$ is smaller than 1, we can write
\begin{equation}
\label{eq3:err}
(I+\alpha G^{-1}\tilde{L})^{-1}=I - \alpha G^{-1}\tilde{L}+\alpha^2 G^{-2}\tilde{L}^2-\alpha^3 G^{-3}\tilde{L}^3+\ldots
\end{equation}
and
\begin{align*}
\label{eq4:err}
&\Vert (I+\alpha G^{-1}\tilde{L})^{-1} - (I+\alpha G^{-1}{L})^{-1}\Vert  \\
&\leq \alpha \Vert 
 G^{-1}(\tilde{L}-L)\Vert +\alpha \Vert G^{-2}(\tilde{L}^2 -L^2)\Vert + \alpha^2 \Vert G^{-3}(\tilde{L}^3-L^3)\Vert +\ldots
\end{align*}
In general, the assumption $\Vert W - \tilde{W} \Vert \leq \varepsilon  $ is not sufficient to say something about the error $\Vert ({W}^{-1}- \tilde{W}^{-1}\Vert $ because the later is proportional to the condition number of $\tilde{W} $, which could be very large. The reason for a large condition number is that the smallest eigenvalue could lie very close to zero. In this case some regularization may help (e.g., adding a positive number to all diagonal elements, similar to Tikhonov regularization). In this sense, the diagonal matrix $G$ helps to bound the error $\Vert ({G}+\alpha \tilde{L})^{-1}- ({G}+\alpha {L})^{-1}\Vert $.
We remind that by one of the properties of the graph Laplacian states $\det(L)=0$ and $L$ is not invertible.}
%
Assume now that instead of Eq.~\ref{eq:defH} we have an $\H$-matrix approximation
$\tilde{H}$ of $H$.
Then the $\H$-matrix approximation of the graph Laplacian will be $\tilde{L}'=\tilde{D}'-\tilde{H}$, where
$\tilde{D}'=\text{diag}(\tilde{D}'_{11},$ $\dots, \tilde{D}'_{nn})$, $\tilde{D}'_{ii} =
\sum\limits_{j} \tilde{H}(i,j)$. It is important to notice that the computational cost of computing $\tilde{D}$ is $\mathcal{O}(kn\log n)$, $k\ll n$.

Substituting $\tilde{L}'$ in (\ref{Hf*}), we obtain cluster ensemble based
predictions of output feature in semi-supervised regression:
\begin{equation} \label{Hf**}
\tilde{f}^{**}=({G}+\alpha \tilde{L}')^{-1}\;Y_{1,0}.
\end{equation}
Here we cannot apply the Woodbury formula, but we also do not need it since the computational cost of computing $({G}+\alpha \tilde{L}')^{-1}$ in the $\H$-matrix format is just $\mathcal{O}(k^2n\log^2 n)$.


The SSR-LRCM Algorithm requires only minor changes, namely, in the second step we compute an $\H$-matrix representation of the graph Laplacian and on the third step calculate predictions of target feature according to
(\ref{Hf**}). The total computational complexity is log-linear.

\section{Numerical experiments}
In this section we describe numerical experiments with the proposed
SSR-LRCM algorithm. The aim of experiments is to confirm the
usefulness of involving cluster ensemble for similarity matrix
estimation in semi-supervised regression. We experimentally evaluate
the regression quality on a synthetic and a real-life example.

\subsection{First example with two clusters and artificial noisy data }
In the first example we consider datasets generated from a mixture
of two multidimensional normal distributions
$\mathcal{N}(a_1,\sigma_X I)$, $\mathcal{N}(a_2,\sigma_X I)$ under
equal wei\-ghts; $a_1$, $a_2$ $\in \mathbb{R}^d$, $d=8$, $\sigma_X$
is a parameter. Usually such type of data is applied for a
classifier evaluation; however it is possible to introduce a real
valued attribute $Y$ as a predicted feature and use it in regression
analysis. Let $Y$ equal $1+\varepsilon$ for points generated from
the first distribution component, otherwise $Y=2+\varepsilon$, where
$\varepsilon$ is a Gaussian random value with zero mean and variance
$\sigma^2_\varepsilon$. To study the robustness of the algorithm, we
also generate two independent random variables following uniform
distribution $\mathcal{U}(0,\sigma_X)$ and use them as additional
``noisy'' features.

In Monte Carlo modeling, we repeatedly generate samples of size $n$
according to the given distribution mixture. In the experiment, 10\%
of the points selected at random from each component compose the
labeled sample; the remaining ones are included in the unlabeled
part. To study the behavior of the algorithm in the presence of
noise, we also vary parameter $\sigma_\varepsilon$ for the target
feature.

In SSR-LRCM, we use $K$-means as a base clustering algorithm. The
ensemble variants are designed by random initialization of centroids
(number of clusters equals two). The ensemble size is $r=10$. The
wights of ensemble elements are the same: $w_l\equiv 1/r$. The
regularization parameters $\alpha, \beta$ have been estimated using
grid search and cross-validation technique. In our experiments, the
best results have been obtained for $\alpha=1$, $\beta=0.001$, and $\sigma_X=5$.

For the comparison purposes, we consider the method (denoted as
SSR-RBF) which uses the standard similarity matrix evaluated with RBF
kernel. {Different values of parameter $\ell$ were considered and the quasi-optimal $\ell= 4.47$ was taken}. The output predictions are
calculated according to formula (\ref{f*}).

The quality of prediction is estimated as Root Mean Squared Error:
$\text{RMSE}=\sqrt{\frac{1}{n}\sum (y_i^{\text{true}}-f_i)^2}$,
where $y_i^{\text{true}}$ is a true value of response feature
specified by the correspondent component. To make the results more
statistically sound, we have averaged error estimates over 40 Monte
Carlo repetitions and compare the results by paired two sample
Student's t-test.

Table \ref{T1} presents the results of experiments. In addition to
averaged errors, the table shows averaged execution times for the
algorithms (working on dual-core Intel Core i5 processor with a
clock frequency of 2.8 GHz and 4 GB RAM). For SSR-LRCM, we
separately indicate ensemble generation time $t_{\text{ens}}$ and
low-rank matrix operation time $t_{\text{matr}}$ (in seconds). The
obtained $p$-values for Student's $t$-test are also taken into
account. A $p$-value less than the given significance level (e.g.,
$0.05$) indicates a statistically significant difference between the
performance estimates.

\begin{table} \centering
\caption{Results of experiments with a mixture of two distributions.
Significantly different RMSE values ($p$-value $<10^{-5}$) are in
bold. For $n=10^5$ and $n=10^6$, SSR-RBF failed due to unacceptable
memory demands.}\label{T1}
\begin{center}
\renewcommand{\arraystretch}{1.3} 
\begin{tabular}{|c|c|c|c|c|c|c|}  \hline
  \multirow{2}{36pt}{\hskip12pt $n$} &\multirow{2}{24pt}{$\;\;\sigma_\varepsilon$}
  & \multicolumn{3}{|c|}{SSR-LRCM} & \multicolumn{2}{c|}{SSR-RBF}
  \\
  \cline{3-7} & & \; RMSE \; & $\;t_{\text{ens }}$ (sec)\;
  & \;$t_{\text{matr }}$  (sec) \;
  & \;RMSE\; & \;  time  (sec)\;  \\
\hline \multirow{3}{36pt}{\;\;1000}
 & 0.01 & \textbf{0.052} & 0.06 & 0.02 & \textbf{0.085} &  0.10  \\
 & 0.1 & \textbf{0.054} & 0.04 & 0.04 & \textbf{0.085} &  0.07  \\
 & 0.25 & \textbf{0.060} & 0.04 & 0.04 & \textbf{0.102} &  0.07  \\
 \hline \multirow{3}{36pt}{\;\;3000}
 & 0.01 & \textbf{0.049} & 0.06 & 0.02 & \textbf{0.145} &  0.74  \\
 & 0.1 & \textbf{0.051} & 0.06 & 0.02 & \textbf{0.143} &  0.75 \\
& 0.25 & \textbf{0.053} & 0.07 & 0.02 & \textbf{0.150} &  0.79 \\
\hline \multirow{3}{36pt}{\;\;7000}
 & 0.01 & \textbf{0.050} & 0.16 & 0.08 & \textbf{0.228} & 5.70  \\
 & 0.1 & \textbf{0.050} & 0.16 & 0.08 & \textbf{0.229} &  5.63  \\
& 0.25 & \textbf{0.051} & 0.14 & 0.07 & \textbf{0.227} &  5.66  \\
 \hline \multirow{1}{36pt}{\;\;\;$10^5$}
 & 0.01 & 0.051 & 1.51 & 0.50 & - &  -  \\
  \hline \multirow{1}{36pt}{\;\;\;$10^6$}
 & 0.01 & 0.051 & 17.7 & 6.68 & - &  -  \\
 \hline
\end{tabular}
\end{center}
\end{table}
The results show that the proposed SSR-LRCM algorithm has
significantly smaller prediction error than SSR-RBF. At the same
time,  SSR-LRCM has run much faster, especially for medium sample
size. For a large volume of data ($n=10^5$, $n=10^6$) only SSR-LRCM
has been able to find a solution, whereas SSR-RBF has refused to
work due to unacceptable memory demands (74.5GB and 7450.6GB
correspondingly).
%
\subsection{Second example with 10-dimensional real Forest Fires dataset }

In the second example, we consider Forest Fires dataset \cite{https2}.
It is necessary to predict the burned area of forest fires, in the
northeast region of Portugal, by using meteorological and other
information. Fire Weather Index (FWI) System is applied to get
feature values. FWI System is based on consecutive daily
observations of temperature, relative humidity, wind speed, and
24-hour rainfall. We use the following numerical features:
\begin{itemize}
    \item X-axis spatial coordinate within the Montesinho park map;
    \item Y-axis spatial coordinate within the Montesinho park map;
    \item Fine Fuel Moisture Code;
    \item Duff Moisture Code;
    \item Initial Spread Index;
    \item Drought Code;
    \item temperature in Celsius degrees;
    \item relative humidity;
    \item wind speed in km/h;
    \item outside rain in mm/m2;
    \item the burned area of the forest in ha (predicted feature).
\end{itemize}

This problem is known as a difficult regression task \cite{Cortez},
in which the best RMSE was attained by the naive mean predictor. We
use quantile regression approach: the transformed quartile value of
response feature should be predicted.

The following experiment's settings are used. The volume of labeled
sample is 10\% of overall data; the cluster ensemble architecture is
the same as in the previous example. $K$-means base algorithm with
10 clusters with ensemble size $r=10$ is used. Other parameters are $\alpha=1$,
$\beta=0.001$, the SSR-RBF parameter is $\ell = 0.1$. The number of
generations of the labeled samples is 40.

As a result of modeling, the averaged error rate for SSR-LRCM has
been evaluated as RMSE$=1.65$. For SSR-RBF, the averaged RMSE is equal to $1.68$.
The $p$-value which equals $0.001$ can be interpreted as indicating
the statistically significant difference between the quality estimates.
\section*{Conclusion}
In this work, we solved the regression problem to forecast the unknown value $Y$.
For this we have introduced a semi-supervised regression method
SSR-LRCM based on cluster ensemble and low-rank co-association
matrix decomposition. We used a scheme of a single clustering
algorithm which obtains base partitions with random initialization.

The proposed method combines graph Laplacian regularization and
cluster ensemble methodologies. Low-rank or hierarchical decomposition of the
co-association matrix gives us a possibility to speedup calculations
and save memory from cubic to (log-)linear.

There are a number of arguments for the usefulness of ensemble
clustering methodology. The preliminary ensemble clustering allows
one to restore more accurately metric relations between objects
under noise distortions and the existence of complex data structures.
The obtained similarity matrix depends on the outputs of clustering
algorithms and is less noise-addicted than the conventional
similarity matrices (eg., based on Euclidean distance). Clustering
with a sufficiently large number of clusters can be viewed as Learning
Vector Quantization known for lowering the average distortion in
data.

The efficiency of the suggested SSR-LRCM algorithm was confirmed
experimentally. Monte Carlo experiments have demonstrated
statistically significant improvement of regression quality and
decreasing in running time for SSR-LRCM in comparison with analogous
SSR-RBF algorithm based on standard similarity matrix.

In future works, we plan to continue studying theoretical properties
and performance characteristics of the proposed method. Development
of iterative methods for graph Laplacian regularization is another
interesting direction, especially in large-scale machine learning
problems. We will further research theoretical and numerical properties 
of the $\H$-matrix approximation of $W$ and $H$.
Applications of the method in various fields are also
planned, especially for spacial data processing and analysis of
genetic sequences.

\section*{Acknowledgements}

The work was carried on according to the scientific research program
``Mathematical methods of pattern recognition and prediction'' in
The Sobolev Institute of Mathematics SB RAS. The research was partly
supported by RFBR grants 18-07-00600, 18-29-0904mk, 19-29-01175 and partly by the
Russian Ministry of Science and Education under the 5-100 Excellence
Programme. A. Litvinenko was supported by funding from the Alexander von Humboldt Foundation.


\end{document}